\documentclass[preprint]{acm_proc_article-sp}

\usepackage{color}

\usepackage{tabularx}
\usepackage{multirow}
\usepackage{amsmath}
\usepackage{hhline}
\usepackage{authblk}
\usepackage{algorithm}
\usepackage[noend]{algpseudocode}
\usepackage{placeins}

\usepackage{url}

\begin{document}

\title{Interpreting social cues to generate credible affective reactions of virtual job interviewers}

\author[1]{\bf {\large Haza\"el Jones}}
\author[2]{\bf {\large Nicolas Sabouret}}
\author[3]{\bf {\large Ionut Damian}}
\author[3]{\bf {\large Tobias Baur}}
\author[3]{\bf {\large Elisabeth Andr\'e}}
\author[4]{\bf {\large Ka\'ska Porayska-Pomsta}}
\author[4]{\bf {\large Paola Rizzo}}

\affil[1]{Laboratoire d'Informatique de Paris 6,4 place Jussieu, 75005 Paris, France, \textit {hazael.jones@lip6.fr}}
\affil[2]{Laboratoire d'Informatique pour la M\'ecanique et les Sciences de l'Ing\'enieur, 91403 Orsay, France, \textit {nicolas.sabouret@limsi.fr}}
\affil[3]{Human Centered Multimedia, Augsburg University, 86159 Augsburg, Germany, \textit {\{damian, baur, andre\}@hcm-lab.de}}
\affil[4]{London Knowledge Lab, Institute of Education, London, UK, \textit {\{K.Porayska-Pomsta, P.Rizzo\}@ioe.ac.uk}}

\renewcommand\Authands{ and }
\maketitle
\begin{abstract}
In this paper we describe a mechanism of generating credible affective reactions in a virtual recruiter during an interaction with a user. This is done using communicative performance computation based on the behaviours of the user as detected by a recognition module. The proposed software pipeline is part of the TARDIS system which aims to aid young job seekers in acquiring job interview related social skills. In this context, our system enables the virtual recruiter to realistically adapt and react to the user in real-time.



\end{abstract}




\section{Introduction}

The number of young people who are not in employment, education or training (NEETs) is increasing
across Europe. According to Eurostat \footnote{\url{ec.europa.eu/eurostat}}, in
March 2012, 5.5 million of 16 to 25 years old European youngsters
were unemployed, amounting to $22.6\%$ of the youngster global
population, which is 10 points above the entire world's population.  These statistics highlight European youth unemployment as a significant problem.

NEETs often lack self-confidence and the essential social skills needed to seek and secure employment \cite{Bynner2002}. The young unemployed often find it difficult to present themselves in a good light to prospective employers, which may put them at further risk of marginalisation. Social coaching workshops, organized by youth inclusion associations across Europe, constitute a common approach to helping people in acquiring and improving their social competencies, especially in the context of job interviews. However, this is an expensive and time-consuming approach that relies on the availability of trained practitioners as well as a willingness of the young people to discuss their strengths and weaknesses in front of practitioners  and often also in front of their peers.

{\bf Digital games} offer a promising way of supporting the training and coaching of young people, providing them with a safe and private environment in which they can practice their skills repeatedly.  TARDIS\footnote{\url{http://www.tardis-project.eu}} \cite{tardis2013} is a project funded by the FP7, whose aim is to build a scenario-based serious-game simulation platform that supports social training and coaching in the context of job interviews. It relies on the use of virtual agents that are capable of reacting in real-time to a human interlocutor's affect, based on the interlocutor's non-verbal cues that are automatically detected and analysed in the context of simulated job interviews.

The goal of this paper is to present the TARDIS' mechanisms involved in the detection and interpretation of social cues and in informing the behaviours of virtual recruiters. We show how the system builds affective states and beliefs about a real user during an interaction with a virtual character. Specifically, the TARDIS game relies on real-time social cue recognition, communicative performance computation and affective computation/decision making by the virtual recruiter.  Building a credible job interview simulation involving a reactive virtual agent requires real-time information about the youngster in order to allow for an assessment of the appropriateness of the youngster's reactions and their communicative performance. Such assessment is done through the perception of a number of relevant social cues that are interpreted in term of performance, relative to the social expectations that are associated with a given situation. This assessment allows the TARDIS system to compute both an affective reaction for the virtual recruiter and the future steps in the interaction dialogue.

The next section presents a brief overview of related work. We then present the different components of our architecture that have been implemented in the TARDIS system.

\section{State of the art} \label{StateOfArt} 

Several research projects have already considered using virtual agents to help humans improve their social skills and, more generally, their emotional intelligence \cite{goleman2006social,tartaro2008playing,aylett2009affective}. However, while the existing approaches use a reactive approach to the user inputs, our efforts focus on deriving an affective reaction of the virtual recruiter, based on the analysis of the user's beliefs and social cues. 

\subsection{Social Cue Recognition}

Using signal processing techniques to detect behavioural patterns is not a new idea (for an overview see \cite{vinciarelli12}). However, most research to date focused on a reduced number of modalities, such as speech \cite{vogt05} or facial expressions \cite{zeng09}, to infer user states, with little attention having been paid to gestures or postures \cite{kapoor05,kleinsmith11}. Furthermore, signal processing work is aimed for offline analysis of recordings, rather than real-time interactive applications. One interesting example is the public training system presented by Batrinca et al. \cite{batrinca2013}, where the user is able to practice public speaking with the help of a virtual crowd. However, the behaviour analysis that they propose does not happen in real-time during the interaction, but rather post-hoc and offline. This means that the system is not able to react to the user's behaviour as they interact with the system.

There are, of course, exceptions. One recent example is the job interview simulation system presented by Hoque et al. \cite{hoque2013mach}. Their system is able to detect a limited number of social cues in real-time, including smiles and audio features as well as perform speech recognition. While, owing to the large number of languages that TARDIS aims to support, our system does not implement speech recognition, it does recognise a much broader range of social cues, including bodily social cues, such as movement energy, gestures and postures, physiological features and eye gaze.

\subsection{Job interview assessment based on social cues}
Several researchers show that diverse interaction scenarios, including  job interviews, negotiations or group meetings, are heavily impacted by the interactants' nonverbal behaviours. For instance \cite{McGovern1979} and \cite{Arvey1982} found that the assessment of candidates by job interviewers is significantly influenced by social cues such as tone of voice, eye gaze contact and body movement. For these reasons, the contemporary research in this field increasingly focuses on the relationship between non-verbal behaviours and the outcome of interactions. Specifically to job interviews, \cite{Curhan2007} studied how the success of simulated job interviews can also be predicted by conversational engagement, vocal mirroring, speech activity, and prosodic emphasis. 

The training of social skills has informed several computer-based simulation environments for domains that include bullying at school \cite{aylett2009affective}, intercultural communication \cite{kistler12} and job interviews \cite{prendinger:ishizuka:2005}. However, in the case of job interview situations a major difficulty is that the interviewees tend to suppress their emotions and  avoid showing social cues, which may result in an undesirable impression being given to the interviewer and ultimately it may impact negatively on the interview outcome. Emotional expression in job interviews is heavily governed by display rules, and most job applicants feel obliged to hide their feelings, especially if those feelings are negative \cite{Sieverding2009}. The fact that in job interview situations, young job seekers tend to avoid showing emotions has been confirmed by our own studies, which will be reported in future publications.  In these studies,  subjects involved in both mock interviews with human recruiters and in simulated job interviews with virtual recruiters displayed very few or no bodily social cues that could be detected reliably and consistently by our current technologies. For this reason, we are currently focusing on voice cues.

\subsection{Affects and Theory of Mind}
Several approaches to building credible virtual humans have been proposed in the domain of affective computing\footnote{See the Humaine project: \url{emotion-research.net}}, including cognitive models of emotions \cite{Ortony1988b,Marsella2003}, models of personality \cite{Prendinger2001} and of social relations \cite{Ochs2010}. However, to our best knowledge, no computational model of social attitudes has been proposed. Social attitudes are socially conditioned expressions of the personality of an agent as manifested in its behaviours and emotional displays. The social conditioning derives from the specific social and cultural norms of a given community or situational contexts. For example, in the context of a job interview, social attitudes provide the recruiter with information about the interviewee's personality and feelings towards the job. This information influences the way in which the virtual recruiter will progress the interview and how it will perceive the suitability of the interviewee for the given job.  Many of the processes involved in interpreting the beliefs and emotions of the TARDIS' users are cognate with and draw from the studies related to Theory of Mind \cite{Meltzoff1995} and reverse appraisal \cite{Hareli2010}.

Theory of mind (ToM) \cite{Leslie1994,BaronCohen95}, is the ability of a person to attribute mental states (beliefs, intentions, desires and affects) to others.  Numerous studies have been conducted in relation to the reasoning process of an agent about the cognitive process of another agent \cite{Bosse07, Dastani2012}. In our work, we want to model the reasoning process of an agent that deduces  the preferences of an human (the interviewee). This particular configuration raises additional difficulties and leads to a novel formulation of the ToM. As a way of illustrating the novelty of our approach, consider for example Pynadath's approach \cite{Pynadath2005}, where an agent has subjective beliefs about others. Artificial agent A has beliefs about another artificial agent B, which beliefs follow the real structure of agent B's beliefs. However, since in TARDIS, agent B is human (the interviewee), we do not have access to prior access to its belief structure, which must be inferred from the outputs of the affect recognition module.

\section{Architecture overview} \label{Architecture}

\begin{figure*}[h!t]
\begin{center}
\includegraphics[width=0.85\linewidth]{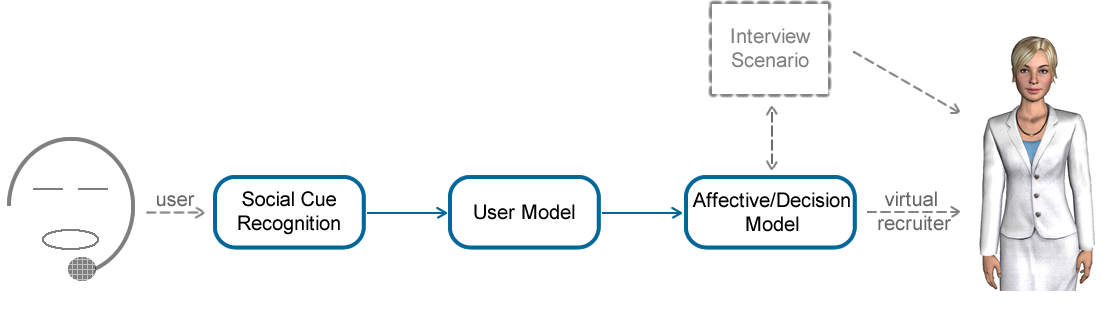}
\caption{Global architecture}
\label{fig_glob_archi}
\end{center}
\end{figure*}

The TARDIS architecture is composed of five main components as shown in Figure \ref{fig_glob_archi}.

\paragraph{The Social Cue Recognition (SCR)} This module detects and recognises various social cues that are produced by the youngsters during the simulated interviews. The module sends the detected cues to the on-line user model. To facilitate detection, the module requires the use of different sensors, such as microphones or cameras. In this paper, we will focus on the audio cues perceived with the help of a head-mounted microphone.

\paragraph{The Online User Model} This module receives information about the youngster's social cues detected by the SCR module, it computes a performance index based on the average values of audio social cues that can be considered suitable to a job interview, and it sends this information to the Affective Core, described in Section \ref{affectivecore}.

\paragraph{The Interview Scenario} This component provides the virtual recruiter with the expectations that it should have in relation to the interviewed youngster's emotions and attitudes, based on the specific stage in the interview. In the current version of TARDIS, the agent has no understanding of the youngster's actual answers to the questions. It simply follows a predetermined scenario, while focusing on the affective recognition and adaptation, based on its knowledge of the scenario and the specific interview questions that have been asked at any given point. It is well known in the virtual character research community that the definition of context-sensitive interactive scenario often requires expert programmers and typically results in a hard-to-maintain code. To avoid this, a visual authoring approach for virtual character applications is used. This tool is based on an existing authoring tool, the Visual SceneMaker that allows the scenario to be built by non computer programmers. The scenario provides a context to the virtual recruiter model, specifically -- the expectations of the recruiter with respect to the desirable emotions and attitudes of the youngsters. 

\paragraph{The Affective and Decision Model of the Virtual Recruiter} The affective and decision model represents the beliefs of the virtual recruiter about the youngster's mental states (specified in terms of affect). It also represents the recruiters intentions with respect to its possible best next interview action. The affective core periodically computes the new affective states for the Virtual Recruiter, based on the perceptions and scenario-based expectations, and current affective states of the youngsters. The difference between the expected and the actually expressed affective states of the youngster are key in the update of the virtual recruiter's own affective states (see next section).  Based on the system's theory of mind of the youngster, the decision module selects the next actions in the scenario.

\paragraph{The Virtual Recruiter Animation} The animation module is responsible for rendering the virtual recruiter's affective state through its behaviours. To achieve realistic behaviour displays, we developed a new motion capture pipeline able to process the data captured more efficiently. The expression of affects relies on a model of interpersonal stance, which in turn is based on different time windows. This new pipeline is integrated with Greta\footnote{GRETA is an Embodied Conversational Agent used in the SEMAINE project.} and it uses MPEG-4, BML and FML standards to allow the user to define the agent's behaviours in XML like form. This allowed us the easy integration of the Charamel\footnote{\url{http://charamel.de/}} agent Gloria. 

In order to determine the social cues and affects that are relevant during job interviews, a knowledge elicitation study was conducted with ten users in France\footnote{The study was conducted at Mission Locale, which is a French network of organisation that helps youngster in their social course and professional career.}. These experiments involved enactments of job interviews as they are normally conducted in the participating organisation. The studies allowed us to determine the affective states of the youngsters that are relevant to this domain of interaction and that are integrated in our youngster and recruiter models.  Based on the knowledge elicited from these studies, the TARDIS' dialogue model has also been designed.

The rest of this paper focuses on three important modules in the TARDIS simulation. In Section \ref{socialcues}, we present the social cue recognition module, Section \ref{usermodel} focuses on the communicative performance computation, while the recruiter affective and decisional model is discussed in Section \ref{affectivecore}.
\section{Social cue recognition} \label{socialcues}

The Social Cue Recognition module is based on the Social Signal Interpretation framework \cite{SSI2013}. SSI provides an interface to diverse sensing devices, as well as a variety of tools for real-time recording and pre-processing of data of human behaviours. The functionality of the module regarding the recognition of body social cues, such as gestures, postures or facial expressions, has been presented in our previous work \cite{damian2013,baur2013socialcom,baur2013nova}. In this paper, we will focus on the detection, analysis and interpretation of the audio cues. Table \ref{tab:socialcues} summarizes the relevant cues.

\begin{table}[h!t]
\begin{center}
\begin{tabular}{|p{3cm}|p{4.5cm}|} \hline
\textbf{Social Cue} & \textbf{Description} \\ \hline
voice activity & Presence or absence of voice \\ \hline
intensity, loudness, energy & Energy-based features of the audio signal\\ \hline
pitch value & The pitch (F0) of the audio signal \\ \hline
jitter, shimmer, voice breaks, harmonicity & Quality-of-voice features \cite{praat} computed from pitch \\ \hline
speech rate & Rate of user's speech \cite{praat} \\ \hline
length of speech segments & The duration in seconds of the user's speech segments determined by voice activity detection \\ \hline
\end{tabular}
\end{center}
\caption{Audio cues recognised by the SCR module}
\label{tab:socialcues}
\end{table}

%
%
%


As a first step in the recognition of social cues, the raw data coming from the sensors is filtered and transformed with the help of an SSI pipeline and various third party libraries, such as PRAAT \cite{praat} and OpenSMILE \cite{eyben2010}. The data streams are then forwarded to the individual social cue recognisers which trigger social cue occurrence events.



The SCR module is able to detect two types of social cues: discrete and continuous. In the case of discrete social cues, the recogniser triggers an event as soon a sufficiently strong occurrence of the respective social cue has been detected. Assessing the quality of a given social cue is either performed using a threshold approach or machine learning algorithms. The event contains information on the recognised social cue and the time of the occurrence, which is meant to be processed by the online user model that makes inferences about the user's complex mental states.

Continuous social cues are sent once per second and contain information on the current magnitude or probability of the observed social signal.

\section{Using audio cues to compute \\ communicative performance} \label{usermodel}

The online user model, currently being implemented, compares the audio cues data produced by users with a set of expected or desirable cues, i.e. audio features that are deemed suitable for each part of a job interview. Based on this comparison, the model computes an index between the values of $-1$ (negative performance) and $+1$ (positive performance), and sends it to the Affective Core. $0$ represents neutral performance. The index is a weighted sum of the differences between each desired cue value and the actual value produced by the youngster regarding speech duration, speech rate, speech volume, pause before speech, etc.

The  following are examples of expected cues:
\begin{itemize}
\item interviewer's questions that are complex or sensitive in nature, e.g.  ''do you have any weaknesses?" requires an answer that goes beyond a simple ''yes'' or a ''no''.  An appropriate answer should be elaborate and thus, a short speech duration in response to a complex question such as exemplified, would be assigned a performance index close to $-1$;
\item the voice should not be too loud (intense), as it might sound aggressive, nor too quiet, as it might  give the impression of lacking self-confidence;
\item the speech should not be too fast, because it might be considered a sign of anxiety, nor too slow, as it might indicate boredom or disengagement;
\item the interviewee should not begin to answer too long after the interviewer's questions, because this could be deemed a sign of hesitation, but equally, it the interviewee should not start answering before the interviewer finishes their question;
\item the voice should have a good variability of pitch, so as not to sound boring.
\end{itemize}

Our studies of human-to-human mock job interviews have shown that there are strong individual differences among youngsters' social cues. For this reason the on-line user model requires a calibration phase to take place during an initial part of the interview (over at least the first 3 turns), and then it continuously compares the youngster's behaviours to their typical baseline as established during calibration, it identifies peaks (i.e. behaviours significantly above or below the baseline), and compares them with the expected cues in order to compute the communicative performance index.

\section{Affective \& Decisional Core} \label{affectivecore}

The perception module and the scenario module provide the input to this module, which is a performance index of the youngster in the range of $[-1, +1]$. This index is used to compute recruiter affects and decision.


\subsection{Affective Core}

\subsubsection{Overview of the model}
The Affective Module is based on a set of rules that compute categories
of emotions, moods and attitudes for the virtual recruiter, based
on the contextual information given by the scenario and the detected
affects (emotions, moods and attitudes) of the participant.  The computation of the virtual agent's emotions is based on the OCC model \cite{Ortony1988b} and the computation of the agent's moods is based on the ALMA model \cite{Gebhard2005}. The details of the computation of emotions
and moods will not be presented in this paper; it can be
found in \cite{jones2013}.

The Affective Core receives a performance index as its input.  The performance index, which fall in the range of  $[-1, +1]$, represents the overall performance of the youngster (its attitude, its affects, its vocal performance). The detected ($d$) performance is denoted as $P_d$. Similarly, a set of expected ($e$) performance index is received from the scenario module. This expected performance index is linked to the difficulty of the question. For example, if a question is easy, e.g. ''Did you find us easily?'', the performance index will be near of $1$.

Formally,in our model, all affects of the recruiter correspond to a value in the interval  of $[0,1]$ and we use $\mathcal{A}$ to denote the set of all affects. The different affects are categorised in terms of three subsets: $E(t)$ (emotions), $M(t)$ (moods) and $A(t)$ (attitudes) are virtual recruiter's simulated affects. These emotions are computed using expert rules based on the values of $\mathcal{P}_d(t)$ and $\mathcal{P}_e(t)$. All these rules are described in \cite{jones2013}.

The list of virtual recruiter's possible affects (emotions, moods and attitudes) that are represented in the model is given in Table~\ref{RecruiterAffects}. Note that this set is different from the
affects that are actually detected and expected. It is based on the literature and on the mock interview corpus analysis (especially the knowledge elicitation phases mentioned in sections \ref{socialcues} and \ref{usermodel}). The emotions are a simple subset of the OCC model \cite{Ortony1988b} that was selected based on what practitioners, acting as recruiters,  expressed during the mock interviews analysed. The moods originated from the ALMA model \cite{Gebhard2005} are defined on 3 dimensions (Pleasure, Arousal and Dominance), but we limited them to the positive dominance zone (since recruiters do not show submissive moods in the context of job interviews). Moods of Table~\ref{RecruiterAffects} are with positive or neutral dominance. 

\begin{table}[h!t]
\begin{center}
\begin{normalsize}
\begin{tabular}{|c|c|c|}
\hhline{~|-|-|}
\multicolumn{1}{c|}{}& Positive  & Negative \\ \hline
 & Joy & Distress \\ \cline{2-3}
Emotions & Relief & Disappointment \\ \cline{2-3}
 & Admiration & Anger \\ \cline{2-3}
 & Hope & Fear \\ \hline

 & Relaxed & Hostile \\ \cline{2-3}
 Moods & Exuberant & Bored \\ \cline{2-3}
 &  & Disdainful \\ \hline

 & Friendly & Aggressive \\ \cline{2-3}
 Attitudes & Supportive &  \\ \cline{2-3}
 &  & Dominant \\ \cline{2-3}
 &  Attentive & Inattentive \\ \cline{2-3}
 &  & Gossip \\ \hline

\end{tabular}
\end{normalsize}
\end{center}
\caption{Recruiter affects}
\label{RecruiterAffects}
\end{table}

The computation of moods is based on emotions following ALMA \cite{Gebhard2005}. In the context of our interview simulation, the period is determined by the number of cycle question/answer. Each answer leads to the computation of a new emotions set and these emotions influence the interviewer's mood. The basis for the calibration is as follows: after five cycles of a specific emotion (anger for example), the virtual recruiter will be in the corresponding mood (hostile). More details about the mood computation can be found in \cite{jones2013}.

The way we compute attitudes follow this principle: an agent can adopt an attitude according to its personality \cite{Snyder1983} or according to its actual mood \cite{Wegener1994}. For example, an agent with a non-aggressive personality may still show an aggressive attitude if its mood becomes very hostile. The mood compensates the personality and vice versa. For this reason, we use a logical-OR as condition on these two dimensions. As a consequence, in our model, the attitude can be triggered by one of these two dimensions. Then, the maximum value (mood or personality) is kept to compute the corresponding attitude, as is classically done in Fuzzy logics.



\subsection{Decisional Core}
Our main objective is to deduce real user beliefs from the \emph{real user/virtual agent} interaction. In this interaction, inputs (affective states of the user) are given by non verbal signals deduced from social signal interpretation. It can be used on different simulations involving the interaction of a human with a virtual agent: teaching, formation, training, amongst other. The common feature of these simulations is the use of questions by the agent. Our model considers the use of questions in order to manage the context of the answers of the person interacting with the TARDIS.

To summarize, our theory of mind model has three main properties:
\begin{itemize}
	\item The theory of mind is about a real person who interacts with the system
	\item It is centred on affective states interpretation of the person in front of the simulation
	\item It uses the context of questions to analyse user responses.
\end{itemize}

\subsubsection{Context management}

Labels are given to the questions/sentences of the virtual character in order to interpret the answer/reaction of the human in terms of beliefs with respect to some topics. A list of topics can be constructed for each specific application. The set of topics $set_{topic}$ contains N topics:  $\{topic_{1}$,~$topic_{2}$,~\dots, $topic_{N}\}$. Each subject is application dependent and based on the domain of the simulation.
A question is defined by $0$ to $n$ topics.

\subsubsection{Building a Beliefs Model}

In order to build beliefs about the human who interacts with the system, we consider the questions/sentences that were just expressed by the virtual agent (identified by labels about topics) and the quality of the answer of the human from an affective point of view (which is obtained by the social cue recognition module) and the performance index $P_d$.
Based on this information, the agent updates its beliefs about the human on a particular subject. We denote the beliefs of the agent about the human as $B_{Human}(topic_i)$ for $i$ in $\{1,$ \dots, $N\}$.

According to the topic(s) raised by the question/remark of the agent, beliefs will be updated. In pursuance of building the beliefs of the human, we consider its answer (perceived via SSI) and based on $P_d$ decide the degree to which the youngster's answer can be categorised as positive ($+1$), negative ($-1$)  or neutral ($0$).





Based on the human's average answer ($P_d$) and the topic tags of the question/remarks just posed/made by the agent, the beliefs can be computed. Updates of each belief is done with the following formula for each topic :

\begin{algorithm}[h!t]
\caption{Beliefs computation}\label{computeBeliefs}
\begin{algorithmic}
\For{$topic_i \in set_{topic}$}
\State $B_{Human}(topic_i) \gets B_{Human}(topic_i) + \alpha \times P_d$
\EndFor
\end{algorithmic}
\end{algorithm}

$\alpha$ is a weight between $0$ and $1$ that can be altered if we want the recruiter beliefs about the human to evolve quickly ($\alpha = 1$) or not ($\alpha$ near of $0$). It can rely on the personality of the agent. An impulsive agent has an $\alpha$ near the value of $1$ and a moderate one close to $0$.


\subsubsection{Desires and goals}

The desires are used to define the strategic intentions of the agent.
Desires are denoted by $D(B_{Human}(topic_i)$, because desires in an interaction are about beliefs of the human on a particular topic. It corresponds to the beliefs that every person process during an interaction according to the reaction of the interlocutor. For instance $D(B_{John}(football)$ is the desire of the agent about John's knowledge in the $football$ topic.


\subsubsection{Dynamics of goals}

 Social attitudes used can be defined on Leary circumplex \cite{Leary1996}. As shown by Leary, attitudes can be separated into two categories: the positive ones (friendly, supportive, etc.) and the negative ones (aggressive, dominant, etc.).





Based on these two types of attitudes, we define the Algorithm~\ref{computeDesire} in order to enable for the desires of the agent to be updated.

\begin{algorithm}[h!t]
\caption{Desires computation}\label{computeDesire}
\begin{algorithmic}
\If {($Attitude \in set(attitude_-))$}
	\For{$B_{Human}(topic) \in set_{topic}$}
		\If {$(P_d < 0)$}
			\State $D(topic) \gets D(topic) + \alpha \times |P_d|$
		\Else
			\State $D(topic) \gets D(topic) - \alpha \times |P_d|$
		\EndIf
	\EndFor
\EndIf
\If {($Attitude \in set(attitude_+))$}
	\For{$B_{Human}(topic) \in set_{topic}$}
		\If {$(P_d < 0)$}
    		\State $D(topic)\gets D(topic) - \alpha \times |P_d|$
		\Else
			\State $D(topic)\gets D(topic) + \alpha \times |P_d|$
		\EndIf
	\EndFor
\EndIf

\end{algorithmic}
\end{algorithm}

This algorithm works as follows: if the agent has a negative attitude, its will intend to select topics with a negative answer for the human. On the other hand, if the agent has a positive attitude, its desires are about topics with a positive answer from the human.

\subsubsection{Goal selection}

Several strategies can be defined to enable the selection of one of the desires in the list of possible desires. The simplest strategy is to select the desire with the maximum value in the available desires.  At any given point in the interview dialogue, every possibilities (topics) will not be suitable for preserving a logical sequence of the conversation (a scenario for instance).

\section{Conclusion}


In this article, we propose a pipeline to compute the affective reactions in a virtual recruiter from user behaviour. To this end, we employ the use of the SSI framework\cite{SSI2013} to recognize user social cue in real time. The social cues are then forwarded to a online user model which uses them to compute a performance index of the user' behaviour. Based on the performance index as well as on the scenario, affects of the virtual recruiter are computed and a ToM approach allows to reason about the user's preferences.

Within the TARDIS training system, which is meant to aid young job seekers in acquiring job interview pertinent social skills, the pipeline will allow the virtual recruiter to react and adapt to the user's behaviour in real time, thus generating credible interaction.


As part of our future work, we aim to conduct studies to evaluate and validate the functionality of the described methodologies. On the technical side, we aim to improve the robustness of the pipeline, for example, by implementing mechanisms to explicitly deal with recognition errors. Furthermore, we plan to extend the online user model to compute other behavioural characteristics as well as to handle more social cues, such as expressivity features, physiological cues or eye gaze.

\section*{Acknowledgment}
The research leading to this paper has received funding from the European Union Information Society and Media Seventh Framework Programme FP7-ICT-2011-7 under grant agreement 288578.

\bibliographystyle{plain}
\bibliography{bib_idgei}

\end{document}